\documentclass{article}

\PassOptionsToPackage{numbers, compress}{natbib}
\usepackage[preprint]{neurips_2024}

\usepackage{times}
\usepackage{latexsym}
\usepackage{inconsolata}

\usepackage[utf8]{inputenc} 
\usepackage[T1]{fontenc}    
\usepackage{hyperref}       
\usepackage{url}            
\usepackage{booktabs}       
\usepackage{amsfonts}       
\usepackage{nicefrac}       
\usepackage{microtype}      
\usepackage{xcolor}         

\usepackage{multirow}
\usepackage{subfigure}

\usepackage[english]{babel}
\usepackage{moresize}
\usepackage{amsmath}
\usepackage{algorithmic}
\usepackage{balance}
\usepackage{comment}
\usepackage{paralist}
\usepackage{bm}
\usepackage{pgfplots}
\usetikzlibrary{pgfplots.dateplot}

\usepackage{flushend}
\usepackage[english]{babel}
\usepackage{graphicx}

\usepackage{amssymb}
\usepackage{url}
\usepackage{bbm}
\usepackage{longtable}
\usepackage{rotating}
\usepackage{mathrsfs}
\usepackage{enumitem}
\usepackage[linesnumbered,algoruled,boxed,lined]{algorithm2e}
\usepackage{adjustbox}
\usepackage{hyperref}
\usepackage{filecontents}

\usepackage{tikz}
\usepackage{footnote}
\usepackage{wrapfig}
\usepackage{pifont}
\usepackage{fontawesome}   

\usepackage[many]{tcolorbox}
\usepackage{fvextra}

\usepackage{tcolorbox}
\usepackage{tikz}
\usetikzlibrary{tikzmark}
\definecolor{myred}{rgb}{0.7, 0.3, 0.0}
\definecolor{myblue}{HTML}{054488}
\definecolor{mygreen}{HTML}{4CAF50}

\definecolor{BoxFrameColor}{RGB}{100,100,100}
\definecolor{BoxBgColor}{RGB}{248,248,248}
\definecolor{TitleTextColor}{RGB}{255,255,255}
\definecolor{BodyTextColor}{RGB}{50,50,50}
\definecolor{TagColor}{RGB}{41,128,185}        

\definecolor{framecolor}{gray}{0.4}   
\definecolor{bgcolor}{gray}{0.98}     
\definecolor{titlecolor}{gray}{1.0}   
\definecolor{tagcolor}{rgb}{0.2,0.5,0.7} 

\makeatletter
\newcommand*\myfontsize{%
  \@setfontsize\myfontsize{8}{8}%
}
\makeatother

\newcommand{\downpct}[1]{\textcolor{green!45!black}{\ensuremath{\downarrow}#1}}

\def\model{PhoneCLI\xspace}

\title{PhoneCLI: From App Interfaces to Callable Commands for Mobile Agents}

\author{
  Yangqin Jiang ~~~
  Lingrui Xu ~~~
  Chao Huang\textsuperscript{}\thanks{Chao Huang is the Corresponding Author.} \\
  The University of Hong Kong \\
  \texttt{\{mrjiangyq99, lingruixu.db, chaohuang75\}@gmail.com} \\
  \faGithub~\textbf{Github Repo:} \textcolor{blue}{\url{https://github.com/HKUDS/PhoneCLI}
}
}

\begin{document}

\maketitle
\setcounter{footnote}{0}

\begin{abstract}
Mobile GUI agents operate through a perception--action loop: at each step they screenshot the device, invoke a vision--language model (VLM), and emit an action. It is slow, costly, and brittle, yet most of what it does is navigation---and everyday navigation is static, ordered, and endlessly repeated. We present \model, which compiles an app's GUI navigation into callable commands, without any app-internal API, runtime instrumentation, or model training. Offline, \model explores a target app from the outside and distills its screens, interactive elements, and navigation edges into a semantically annotated map; each screen yields one deterministic command: a replay sequence that reaches it. Online, the agent selects a command, verifies it before execution, and then executes it deterministically in sub-second time at zero VLM cost; open-ended interaction and every failure of the compiled path fall back to the embedded VLM interpreter, exactly the pure VLM agent, so compilation can only help. On AndroidLab, \model improves the task success rate while reducing steps and token consumption, and it transfers to AndroidWorld's official M3A agent with consistent efficiency gains. What \model compiles is the app's navigation rather than one run, so it serves new tasks, not only repeated ones.
\end{abstract}


\section{Introduction}
\label{sec:intro}

GUI agents for mobile devices largely operate through a perception–action loop: at each step, the agent captures the screen, invokes a vision–language model (VLM) to reason over the visual information, and emits an action at either the index level or in coordinate form. The agent’s competence is acquired through training on GUI corpora and/or via fine-tuning and reinforcement learning~\cite{liu2024autoglm, xu2026mobilerl, li2025mobileuse, zhang2024large}. This paradigm is confronted with three compounding challenges.

\textbf{(\uppercase\expandafter{\romannumeral1}) Closedness.} Desktop computer-using agents can operate via rich programmatic surfaces—\textit{e.g.}, CLI tools, APIs, or even executable code~\cite{song2026coact, yan2025mcpworld, abhyankar2026osworld}---because the desktop ecosystem provides open interfaces. By contrast, mobile apps offer no comparable privilege. They function as black boxes with no client-side, third-party programmatic interface, and even modern mobile frameworks typically only allow composition with system-level commands~\cite{li2026phoneharness}, leaving the internal state and logic of each app inaccessible. Consequently, the agent is restricted to a screenshot-based loop, which is slow (on the order of seconds per step), costly (requiring one API call per step), and brittle, often leading to hallucinated coordinates and navigation failures.

\textbf{(\uppercase\expandafter{\romannumeral2}) Repetitiveness.} Nevertheless, the navigation induced by the loop targets interfaces whose structure is largely known a priori. Everyday mobile app usage is highly repetitive and follows consistent, ordered patterns~\cite{li2020extent, peters2024social}: for instance, opening a settings page, selecting a tab, and locating a specific item. Such behavior typically traces a small set of static paths repeatedly. Unlike free-form interaction, the navigation structure can therefore be enumerated offline. This motivates the core idea of this work: \textbf{pre-build the repeated navigation once and replay it deterministically thereafter.}

\textbf{(\uppercase\expandafter{\romannumeral3}) Generalization.} The alternative route---training---stores capability in the model weights. However, each new app typically requires additional data collection and/or re-training, which exposes a major bottleneck studied in recent work on RL generalization~\cite{gu2026generalization} and on adaptation to previously unseen apps~\cite{guo2026coadapt}. Re-training is both costly and inflexible: a model that has never encountered an app has no direct shortcut for navigating it. Given the scale of modern app stores, with millions of apps, per-app training is infeasible in principle.

Existing remedies either improve the loop from within---by pairing stronger VLMs with grounding and memory scaffolding~\cite{wang2024mobile, jiang2026openphone}---or incur the training cost to sharpen the agent~\cite{xu2026mobilerl}. However, neither approach eliminates the repeated navigation bottleneck itself. We propose \textbf{\model}, which takes a third route: \textbf{compile an app’s GUI navigation into callable commands}. \textbf{Offline}, \model explores each target app once from the outside, without any app-internal API and without instrumentation. It distills the app’s screens, UI elements, and navigation edges into a structured map, and compiles each screen into a deterministic command—\textit{i.e.}, a replayable action sequence that reliably reaches that screen. This offline procedure does not involve training or model updates, so integrating a new app takes only minutes. \textbf{Online}, the agent maps each task to a command, verifies the selection, and replays it deterministically in sub-second time with zero VLM calls. Only open-ended interactions not covered by the command table are handled by the VLM, which serves as a graceful fallback.

Recent work shares parts of this picture. PreAct~\cite{li2026preact} compiles agent trajectories online to replay the same task. UI-KOBE~\cite{chai2026ui} enables runtime following of an explored UI graph. Trajectory-mining approaches recycle explored paths back into the VLM as textual hints~\cite{xie2025gui, sun2025gui}, meaning that reuse still incurs perception costs. \model differs along the decisive axis: it compiles offline, at the granularity of navigation commands, and then executes them deterministically. As a result, the compiled knowledge supports unseen tasks rather than only repeated ones, and its reuse requires no perception at runtime. In short, PreAct compiles what an agent has done, whereas \model compiles what the app is. 
We discuss our contributions are threefold:
\begin{itemize}[leftmargin=*]
    \item \textbf{GUI-to-CLI compilation.} We present a compilation pipeline that transforms a black-box app---lacking any app-internal API, instrumentation, and training---into a structured app map and a catalog of deterministic callable commands, one per screen. This pipeline provides the agent with a programmatic surface that the app itself does not expose, enabling new apps to be integrated within minutes without any model update.
    \item \textbf{Reliable command invocation.} We design an invocation mechanism based on two-phase routing that selects and verifies a command prior to execution. Deterministic replay places the agent in the target screen in sub-second time at zero VLM cost, while any failure falls back gracefully to the embedded VLM interpreter. By removing the most frequently repeated aspect of mobile app usage---navigation---from the VLM loop, \model preserves the performance of its backbone without degradation.
    \item \textbf{Dual-benchmark evidence.} On AndroidLab~\cite{xu2025androidlab}, \model achieves state-of-the-art performance with zero training, while reducing the step and token cost of successful tasks. On AndroidWorld~\cite{rawles2025androidworld}, it transfers to the official agent with measurable efficiency gains. A controlled three-way ablation further attributes the improvements to the map information and deterministic replay, independently.
\end{itemize}
\section{Methodology}
\label{sec:solution}

\begin{figure}[t]
  \centering
  \includegraphics[width=\textwidth]{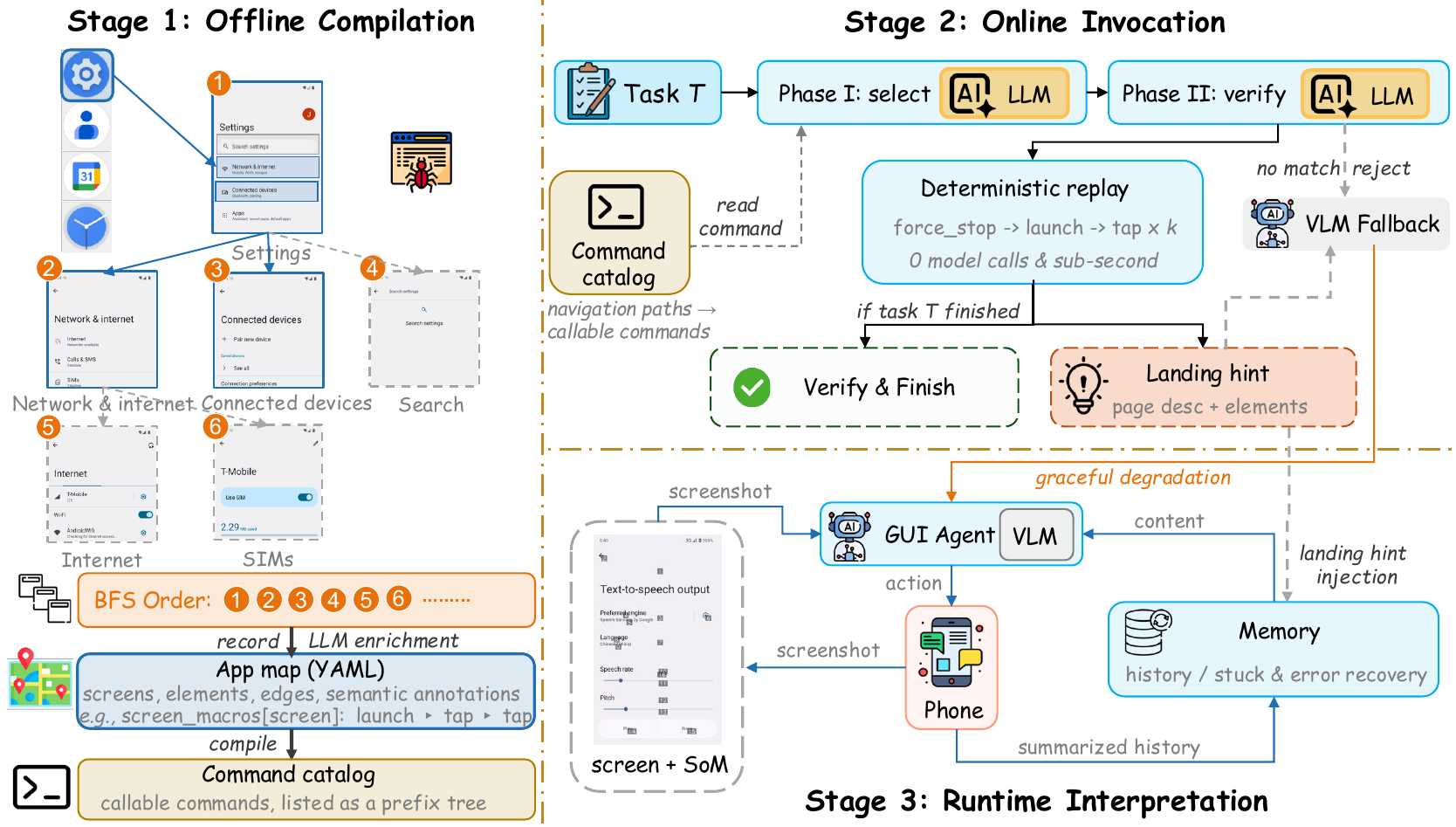}
  \caption{Overall framework of the proposed \model.}
  \label{fig:overview}
\end{figure}

\subsection{Design Overview}
\label{sec:overview}

\textbf{Motivation and design rationale.}
As detailed in Section~\ref{sec:intro}, our design rests on three observations: \textbf{(\uppercase\expandafter{\romannumeral1}) closedness}, mobile apps expose no programmatic surface, unlike the CLI tools and APIs available on desktop, which confines mobile agents to a slow, costly, and brittle screenshot-perception loop; \textbf{(\uppercase\expandafter{\romannumeral2}) repetitiveness}, everyday app usage is highly repetitive and ordered, and the navigation skeleton that connects screens changes far more slowly than app content, so a path distilled once is replayed many times; and \textbf{(\uppercase\expandafter{\romannumeral3}) generalization}, training-based agents are prone to distribution shift on unseen apps and costly to re-train, whereas on-demand exploration integrates a new app in minutes.

This motivates our central design principle: \textbf{build once what is static, and handle on the fly what is not}. Navigation is static, frequent, and pre-enumerable, so we distill it offline into a table of callable commands and reuse it across tasks; open-ended interaction---form filling, dynamic content, unexpected dialogs---is variable and cannot be exhaustively enumerated, so it is handled by the VLM at runtime. Such content---feeds, recommendations, merchant lists---changes on every load; the navigation skeleton does not, and only the skeleton is compiled. We refer to the offline distillation as \textbf{compilation} and to the runtime handling as \textbf{interpretation}---a division of labor analogous to mixed-mode execution in programming languages, where frequently used code is compiled ahead of time and the remainder is interpreted on demand.
Accordingly, \model comprises three stages:
\begin{itemize}[leftmargin=*]
    \item \textbf{Offline compilation (Stage 1, Sec.~\ref{sec:stage1}).} An automated exploration traverses the app from a clean state and compiles its navigation skeleton into a semantically annotated map plus a catalog of commands, turning an interface that can be operated only manually into one that can be invoked programmatically.

    \item \textbf{Online invocation (Stage 2, Sec.~\ref{sec:stage2}).} A router maps a task to a compiled command, the match is confirmed, and the command is replayed deterministically---the analog of invoking a compiled function, with a guard on either side of the call.
    
    \item \textbf{Runtime interpretation (Stage 3, Sec.~\ref{sec:stage3}).} Everything outside the command table is handled by the VLM, which receives every failure of the compiled path as a graceful degradation.
\end{itemize}

\textbf{Formalization.}
Formally, an app map is a structure $\mathcal{G} = (S, E, \lambda, M)$, where $S$ is the set of discovered screens; $E$ is the set of interactive elements, each annotated with normalized coordinates; $\lambda : E \rightharpoonup S \times \mathrm{Alias}^{*}$ is a partial function that records navigation by mapping an element to the screen reached upon its activation together with its semantic aliases, so that a navigation edge is a triple $(s, e, s')$ with $e$ an element of $s$; and $M$ is a set of \emph{commands}, one per screen. Each command $c = [a_1, \dots, a_k]$ is a deterministic action sequence that navigates to its target screen. 

Stage~2 consumes a \emph{catalog} $\mathcal{C}$ derived from $\mathcal{G}$: each entry pairs a stable identifier with the natural language description and target of a command, and carries neither coordinates nor action steps, so that routing operates on symbols rather than on plans. A router $r(T, \mathcal{C}) \mapsto (d, \mathit{id}, \mathit{ans})$ maps a task description $T$ to a decision $d$, the identifier of a selected command when one applies, and an answer string $\mathit{ans}$ when $d = \texttt{FINISH}$; the identifier resolves to a command $c \in M$. The decision takes one of four values: $d = \texttt{OP}$, the command alone completes the task and is replayed with the answerability check; $d = \texttt{MACRO\_VLM}$, the compiled macro performs the required navigation and then transfers control to the VLM, and we call a screen's replay sequence its macro; $d = \texttt{NEED\_VLM}$, no applicable command exists and the VLM explores from the current state; and $d = \texttt{FINISH}$, the task can be answered without device interaction.

\textbf{Design properties.}
\textbf{(\uppercase\expandafter{\romannumeral1}) Compile once, reuse many.} The offline cost is amortized across all subsequent tasks and runs, and navigation no longer consumes the VLM budget. \textbf{(\uppercase\expandafter{\romannumeral2}) Guaranteed floor.} Every failure path---no matching command, a rejected verification, or a landing mismatch---resolves to the pure VLM agent, so the success rate is never below that of the VLM-only baseline; what the compiled path adds is a bounded number of routing calls, not a new failure mode. \textbf{(\uppercase\expandafter{\romannumeral3}) Model-agnostic knowledge acquisition.} The command table is constructed without training or model updates; a new app is integrated in minutes by re-running the offline stage.

\subsection{Offline Compilation: Building App Maps}
\label{sec:stage1}
The first stage constructs the app map $\mathcal{G} = (S, E, \lambda, M)$ from a black-box app in two steps: structural exploration recovers the app's navigation skeleton, and semantic enrichment renders that skeleton interpretable to a language model. Our exploration inherits systematic UI traversal from Android testing crawlers~\cite{su2017guided,gu2019practical}, but targets reusable navigation rather than test coverage. Unlike trajectory-based methods that inject mined knowledge back into the VLM as text~\cite{xie2025gui,sun2025gui}, the explored structure is compiled into deterministic, executable commands.

\textbf{Structural exploration.} Starting from a clean launch state, the crawler traverses the app in breadth-first order. At each screen it reads the UI hierarchy, records the interactive elements it finds, activates them one by one, and observes the screens they reach, thereby assembling the screens, elements, and navigation edges of $\mathcal{G}$. Scrollable screens are explored over multiple scroll pages so that content below the fold is not omitted. The walk is bounded by depth and screen budgets chosen to cover the navigation core of typical apps. Screen deduplication is computed from raw elements before semantic annotation, so that persistent bottom navigation bars shared across tabs do not dominate the signature and collapse distinct tabs into a single screen.

\textbf{Semantic enrichment.} The raw graph captures syntax---text and coordinates---but not semantics: what an element is, or what a screen is for. Because both routing (Stage~2) and verification depend on semantic understanding, a language model annotates the graph. Each element is classified as \texttt{STABLE} (present across app versions) or \texttt{DYNAMIC}; assigned aliases with a semantic type (\texttt{button}, \texttt{input}, \texttt{label}, etc.); and each screen is given a natural language description.


\textbf{Compilation to commands.} From the enriched map, the compiler derives one command per screen: the action sequence that reaches it from a clean launch---typically \texttt{force\_stop}, \texttt{launch}, followed by a sequence of taps and swipes with fixed coordinates and inter-step waits. A command is executable rather than advisory: replaying it reproduces the recorded trajectory exactly, with no model in the loop. Because the map is obtained purely by exploration and LLM annotation, it can also be rebuilt whenever an app changes. An excerpt of the resulting app map is shown in Figure~\ref{fig:map} (Appendix~\ref{app:map}).

\subsection{Online Invocation: Routing and Replaying Commands}
\label{sec:stage2}

Stage~2 is the online counterpart of the compiled command table: given a task, it decides whether a compiled command applies and, if so, executes it reliably. It is deliberately conservative---whenever the mapping is uncertain, the command is declined and the task proceeds to Stage~3.

\textbf{Command routing.} Routing proceeds in two phases, each implemented by a lightweight LLM call. \textbf{Phase I (selection)} presents the task together with a compact command catalog and asks the router to choose among the four decisions defined in Section~\ref{sec:overview}. The catalog is formatted as a prefix tree-commands that share a navigation path are merged at their common prefix and expanded by indentation---so that it remains compact, and it is deliberately symbolic: it carries element paths, semantic tags, and stable operation identifiers, but neither coordinates nor action steps, which keeps it short and prevents the router from inventing device-level actions. A catalog excerpt, together with the router's output format and an example decision, is shown in Figure~\ref{fig:catalog} (Appendix~\ref{app:catalog}). \textbf{Phase II (verification)} inspects the selected command before any execution: the router judges whether the command's target screen, described in natural language, is semantically consistent with the task. If it is not, the command is rejected and the task falls back to the VLM, since replaying a plausible but incorrect command would land the VLM on an irrelevant screen and waste the rounds spent recovering.

\textbf{Deterministic replay and completion.} A selected identifier resolves to the command compiled from the map---the model chooses \emph{what} to do, while the program retains \emph{how}---so replay requires no perception, incurs no model call, and cannot hallucinate coordinates. Execution is therefore deterministic and completes in sub-second time, in contrast to screen-by-screen navigation, where every step costs seconds and one API call. This is where compilation pays off: navigation, the most repeated part of app use, is removed from the VLM loop entirely. A replayed command is then completed in one of two ways. If the command alone solves the task (\texttt{OP}), the agent performs an answerability check on the landed screen and finishes with that answer if it is available; this check differs in kind from the Phase~II verification, which asks whether the command's destination fits the task, whereas this one asks whether the reached screen actually yields the answer. If the answer is not available, the command is demoted to the \texttt{MACRO\_VLM} path, so that a command which navigated correctly but cannot answer on its own still saves the VLM the navigation. To avoid a cold start on an unfamiliar screen, the handoff injects a landing hint: the target screen's natural language description and the list of its interactive elements, together with an explicit note that the command has only navigated to this screen and that the interaction remains to be performed. The hint is cheap---it is drawn from the app map---and turns the apparent teleportation into a warm start.

\textbf{Landing check.} Replay is not trusted blindly. After execution, the current UI hierarchy is matched against the recorded screens; if the agent has demonstrably landed on a screen unrelated to the command's target, the app is restarted to a clean state before the VLM takes over, so that a mismatch never silently misleads it. The command is thus verified before replay and checked afterward, which makes the invocation path self-checking and routes every failure to Stage~3. The full invocation path is summarized in Algorithm~\ref{alg:phonecli}.


\begin{algorithm}[h]
\caption{\model: executing one task}
\label{alg:phonecli}
\small
\SetAlgoLined
\DontPrintSemicolon
\KwIn{Task $T$, app map $\mathcal{G}=(S,E,\lambda,M)$, VLM}
\KwOut{Task outcome (graded success or failure)}
$\mathcal{C} \gets$ catalog of $\mathcal{G}$; \tcp*{Stage 1: symbolic entries, built once per app}
$(d, \mathit{id}, \mathit{ans}) \gets r(T, \mathcal{C})$; \tcp*{Stage 2, Phase I: selection}
\Switch{$d$}{
  \Case{\texttt{FINISH}}{\KwRet{$\mathit{ans}$} \tcp*{answerable without device interaction}}
  \Case{\texttt{NEED\_VLM}}{\KwRet{\textsc{Interpret}(T)} \tcp*{no command applies; continue from the current state}}
  \Case{\texttt{OP} / \texttt{MACRO\_VLM}}{
    $c \gets \textsc{Resolve}(\mathit{id})$; \tcp*{identifier $\to$ recorded action sequence}
    \uIf{$\neg$ \textsc{Verify}($c, T$)}{\KwRet{\textsc{Interpret}(T)} \tcp*{Phase II rejects; current state retained}}
    \textsc{Restart}(app); \textsc{Replay}($c$); \tcp*{clean state; deterministic, zero VLM calls}
    \uIf{$\neg$ \textsc{LandingCheck}($c$)}{\textsc{Restart}(app); \KwRet{\textsc{Interpret}(T)} \tcp*{clean-state fallback}}
    \uIf{$d = \texttt{OP}$ $\land$ answerable on the landed screen}{\KwRet{the answer as $\mathit{ans}$} \tcp*{command alone completes the task}}
    $h \gets$ \textsc{Hint}($c$); \KwRet{\textsc{Interpret}(T, h)} \tcp*{\texttt{OP} demoted, or \texttt{MACRO\_VLM}; Stage 3}
  }
}
\end{algorithm}


\subsection{Runtime Interpretation: VLM Fallback and Graceful Degradation}
\label{sec:stage3}

\textbf{Why interpretation is needed.} Compilation is inherently partial: the map covers only what exploration could observe, and open-ended interaction cannot be enumerated in advance. Stage~3 therefore interprets what was not compiled. The VLM perceives the screen and acts step by step, following the standard perception---action loop of GUI agents~\cite{wang2024mobile,jiang2026openphone}; we write $\textsc{Interpret}(T, h)$ for this loop, where the optional $h$ is the landing hint injected after a replayed navigation.

\textbf{Memory and recovery.} The interpreter maintains a structured state across rounds. Each VLM response ends with a \texttt{STATE\_ASSESSMENT} field, and past assessments are injected into the next prompt in chronological order, giving the agent an explicit memory of what it has tried and where it is stuck. On top of this, the loop detects unproductive behavior---repeated scrolling or waiting without screen progress---and injects targeted hints (\textit{e.g.}, try the search entry; close the ad overlay). When an action fails at execution time, the error is recorded and re-presented in the next round with a corrective hint. These mechanisms turn the interpreter from a stateless screen actor into a small but self-correcting agent.

\textbf{Graceful degradation.} The degradation rule is uniform: whenever the compiled path is unavailable, the system runs the pure VLM agent, continuing from the current state, or from a restarted clean state when the landing check has failed. Stage~3 is thus a co-routine rather than a last resort---it is entered by design on every failure of the compiled path, and the interpreter it supplies is exactly the agent that would otherwise have run alone.
\section{Evaluation}
\label{sec:eval}

\subsection{Experimental Setup}
\label{sec:setup}

\textbf{Benchmarks.} We evaluate on two Android agent benchmarks: AndroidLab~\cite{xu2025androidlab} (138 tasks, 9 apps) and AndroidWorld~\cite{rawles2025androidworld} (116 tasks, 20 apps, open-world with cross-app workflows), where we compare against its official M3A agent.


\textbf{Backbone Models.} We use three backbone models --- Qwen3.7-Plus~\cite{cloud2026qwen37}, GLM-4.6V~\cite{zhipu2026glm46v}, and Kimi-K3~\cite{team2026kimi}. AndroidWorld experiments use Qwen3.7-Plus for both the agent and the baseline, keeping the comparison model-controlled.

\textbf{Baseline methods.} Comparisons use two primary model groups: (1) \emph{General-purpose vision-capable LLMs}: closed-source models (Qwen3.7-Plus~\cite{cloud2026qwen37}, Gemini-2.5-Pro~\cite{comanici2025gemini}, GPT-4o~\cite{hurst2024gpt}, Claude-Sonnet 4) and open-weight models (Kimi-K3~\cite{team2026kimi}, GLM-4.6V~\cite{zhipu2026glm46v}) (2) \emph{GUI-specialized/fine-tuned models}: AutoGLM-Phone, AutoGLM-Mobile~\cite{xu2026mobilerl}, MobileUse~\cite{li2025mobileuse}, UI-Genie-Agent~\cite{xiao2026ui}, UI-Tars-1.5~\cite{qin2025ui}, V-Droid~\cite{dai2025advancing}, and AutoGLM-2024-10~\cite{liu2024autoglm}.
More details of the experimental setup are reported in Appendix~\ref{app:setup}.



\subsection{Overall Performance}
\label{sec:main}

\model is an enhancement layer attached to any GUI agent rather than a standalone backbone, so the first question is whether the attachment itself helps. To answer this, Table~\ref{tab:main} compares, on three backbones, the full \model system against its embedded pure VLM agent (\textbf{VLM-only}) --- \model with the compiled layer removed, all else identical. 

\begin{table}[h]
\caption{Success rate, average steps, and tokens per task on AndroidLab.}
\centering
\small
\begin{tabular}{l l c c l l}
\toprule
Model & Condition & Success & $\Delta$ & Avg. Steps & Avg. Tokens \\
\midrule
\multirow{2}{*}{Kimi-K3} & VLM-only & $68.8\%$ & \multirow{2}{*}{$+0.8$} & $6.34$ & $53.3k$ \\
                         & \model & $\mathbf{69.6\%}$ & & $6.02$$^{\downpct{5\%}}$ & $48.1k$$^{\downpct{10\%}}$ \\
\midrule
\multirow{2}{*}{GLM-4.6V} & VLM-only & $41.3\%$ & \multirow{2}{*}{$+3.2$} & $7.07$ & $61.4k$ \\
                         & \model & $\mathbf{44.5\%}$ & & $6.52$$^{\downpct{8\%}}$ & $54.4k$$^{\downpct{11\%}}$ \\
\midrule
\multirow{2}{*}{Qwen3.7-Plus} & VLM-only & $50.7\%$ & \multirow{2}{*}{$+12.3$} & $7.52$ & $40.1k$ \\
                              & \model & $\mathbf{63.0\%}$ & & $6.72$$^{\downpct{11\%}}$ & $34.4k$$^{\downpct{14\%}}$ \\
\bottomrule
\end{tabular}
\label{tab:main}
\end{table}

Three observations follow: 
\begin{itemize}[leftmargin=*]

    \item \textbf{(i) \model adapts across backbones, with the largest gains on weaker ones.} Attaching \model improves the success rate by $12.3$ points on Qwen3.7-Plus and by $0.8$ points on Kimi-K3. Weaker backbones carry limited app-specific navigation knowledge, and the compiled map supplies exactly this missing knowledge.

    \item \textbf{(ii) Beyond the success-rate gains, tasks are also completed faster and more economically.} On Qwen3.7-Plus, \model reduces the average number of steps by $11\%$ and the token consumption by $14\%$, as deterministic replay replaces multi-round visual navigation with sub-second, zero-token commands.

    \item \textbf{(iii) As backbone capability grows, the success-rate gain shrinks to $\mathbf{0.8}$ points on Kimi-K3, while the cost advantage persists} --- token consumption remains $10\%$ lower and the step count remains $5\%$ lower. An enhancement layer that keeps reducing cost even when it no longer adds capability is particularly valuable under the latency and cost constraints of mobile deployment.

\end{itemize}

\begin{wraptable}{r}{0.45\textwidth}
\caption{Success rate (\%) on AndroidLab.}
\centering
\footnotesize
\begin{tabular}{@{}l c c@{}}
\toprule
Method & Size & SR \\
\midrule
\multicolumn{3}{@{}l}{\textit{General-purpose LLMs}} \\
\midrule
GPT-4o & -- & 31.2 \\
Claude-Sonnet-4 & -- & 40.6 \\
GLM-4.6V & 107B-A12B & 41.3 \\
Qwen3.7-Plus & 35B & 50.7 \\
Gemini-2.5-Pro & -- & 56.5 \\
Kimi-K3 & 2.8T-A104B & 68.8 \\
\midrule
\multicolumn{3}{@{}l}{\textit{GUI-specific models}} \\
\midrule
V-Droid & 8B & 38.4 \\
UI-Tars-1.5 & 72B & 38.4 \\
UI-Genie-Agent & 72B & 41.2 \\
MobileUse & 72B & 44.2 \\
AutoGLM-Mobile & 9B & 46.8 \\
AutoGLM-Phone & 9B & 47.7 \\
AutoGLM-2024-10 & -- & 36.2 \\
\midrule
\multicolumn{3}{@{}l}{\textbf{\textit{\model method}} \textit{w Qwen3.7-Plus}} \\
\midrule
\model \textit{w/o} Replay  & 35B & 55.1 \\
\model & 35B & 63.0 \\
\bottomrule
\end{tabular}
\label{tab:sota}
\end{wraptable}

Table~\ref{tab:sota} situates \model against both general-purpose LLMs and GUI-specific models, and two points stand out. \textbf{First, \model lifts a modest backbone toward frontier level:} Qwen3.7-Plus (35B) moves from 50.7 as a plain VLM agent to 63.0 with \model, closing a 19.2 point gap against the 2.8T-A104B open-weight Kimi-K3 (68.8) to 5.8 points --- and on the Kimi backbone itself, \model still adds a further 0.8 points (69.6, Table~\ref{tab:main}). A 35B model enhanced by \model is thus brought within reach of a 2.8T-class one, demonstrating the strong amplification \model provides for weaker models. Notably, 63.0 is also above every GUI-specific model in Table~\ref{tab:sota}, even though those models are trained for GUI control while \model is not.

\textbf{Second, the two \model rows preview where the gain comes from.} \model w/o replay keeps the app map but removes the replay mechanism: the map is only injected as reference text, and the VLM must navigate screens on its own. Information injection alone lifts Qwen3.7-Plus from 50.7 to 55.1, while adding deterministic replay raises it further to 63.0 --- the replay step contributes the larger share of the gain. This shows that \model's improvement is not merely from exposing map information to the VLM, but substantially from a dedicated design: executing that knowledge deterministically, without VLM involvement. Moreover, App.~\ref{sec:androidworld} reports a generalization test in which our method is attached to the official AndroidWorld agent.

\subsection{Ablation Studies}
\label{sec:ablations}

We conduct ablations at two granularities: (\uppercase\expandafter{\romannumeral1}) how much each of \model’s two core components contributes---the compiled map used as information and deterministic replay---and (\uppercase\expandafter{\romannumeral1}) whether any individual mechanism within these components is dispensable. We study both questions on AndroidLab using Qwen3.7-Plus across nine apps.

\begin{figure}[h]
\centering
\includegraphics[width=0.85\columnwidth]{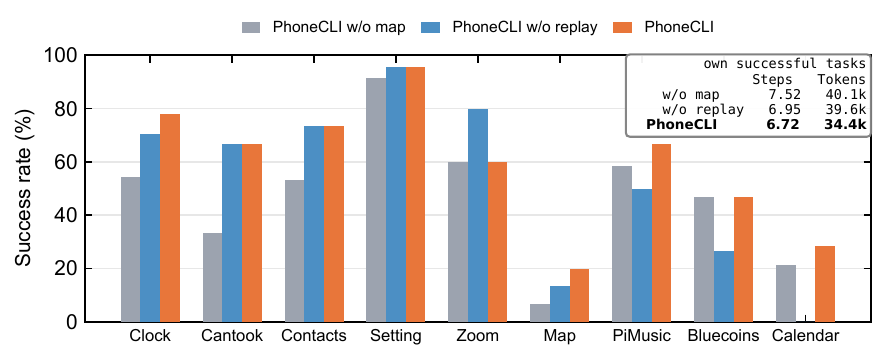}
\caption{Success rate of the three conditions per app on AndroidLab (Qwen3.7-Plus), sorted by the information contribution (left: helpful, right: harmful). Inset: average steps and tokens on each condition's own successful tasks.}
\label{fig:attribution}
\end{figure}

\textbf{(\uppercase\expandafter{\romannumeral1}) Core decomposition.} Three conditions decompose \model's stack along its modules. \emph{w/o map} (VLM only) removes the compiled map together with its replay machinery, leaving Stage~3, the pure VLM agent, at 50.7\%. \emph{w/o replay} keeps the map but only as information, injecting its navigation reference into the VLM's prompt, at 55.1\% ($+4.4$). \emph{\model} additionally executes matched commands deterministically, at 63.0\% ($+7.9$). Both components pay off, replay the more.

\emph{Cost and coupling.} Injection shortens navigation but leaves the VLM in the loop: every remaining round still re-reads the injected reference and still pays a full VLM call, so rounds fall while tokens stay flat (7.52 $\rightarrow$ 6.95; 40.1k $\rightarrow$ 39.6k). Replay removes the call itself --- compiled navigation consumes no VLM call --- and collapses a multi-step traversal into a single recorded operation (6.95 $\rightarrow$ 6.72; 34.4k, $-13\%$). The rounds that remain are the actual interaction rather than navigation. All efficiency numbers average over each condition's own successful tasks (Fig.~\ref{fig:attribution}).

\emph{Failure modes.} Figure~\ref{fig:attribution} orders the nine apps by how much the injected information contributes, and the ordering is uneven: injection helps on some apps and hurts on others. Its failure mode is that every screen must pass through the model's understanding, so it degrades exactly where the map is semantically sparse or noisy --- which is what collapses Calendar to 0/14. Replay has no understanding step and recovers the same app to 4/14, but it executes the command as recorded, so a wrong landing is silent and only a post-replay check can catch it.

\begin{wraptable}{r}{0.48\textwidth}
\vspace{-0.2in}
\caption{Ablation of \model components on AndroidLab (Qwen3.7-Plus,
9 apps).}
\centering
\footnotesize
\begin{tabular}{@{}l c c@{}}
\toprule
Variant & Success & $\Delta$ \\
\midrule
\multicolumn{3}{@{}l}{\textit{Module 1: map annotation}} \\
\midrule
\textit{w/o} element classification & 47.1\% & $-$15.9 \\
\textit{w/o} semantic enrichment & 47.8\% & $-$15.2 \\
\midrule
\multicolumn{3}{@{}l}{\textit{Module 2: routing and replay}} \\
\midrule
\textit{w/o} landing check & 50.7\% & $-$12.3 \\
\textit{w/o} landing hint & 50.0\% & $-$13.0 \\
\textit{w/o} command completion & 48.6\% & $-$14.4 \\
\midrule
\textbf{\model} & \textbf{63.0\%} & -- \\
\bottomrule
\end{tabular}
\label{tab:abl}
\end{wraptable}

\textbf{(\uppercase\expandafter{\romannumeral2}) Fine-grained mechanisms.} Every variant in Table~\ref{tab:abl} removes one mechanism, and every removal costs at least 12 success-rate points. \emph{On the map side}, element classification --- the \texttt{STABLE}/\texttt{DYNAMIC} annotation the router uses to tell version-persistent elements from transient ones --- is the costliest single removal ($-15.9\%$), and semantic enrichment, the aliases and page descriptions that make the structural graph routable to both the router and the verifier, costs nearly as much ($-15.2\%$). \emph{On the replay side}, the landing check verifies that replay actually reached the target screen; without it, wrong landings silently mislead the VLM onto irrelevant screens ($-12.3\%$). The landing hint --- the description and element list injected at handoff --- keeps a teleported VLM from starting cold ($-13.0\%$). Command completion lets the \texttt{OP} path finish a replay-only task in one deterministic replay and one verification, with no VLM interaction at all; forcing such tasks through the VLM wastes that and adds risk ($-14.4\%$).

Across both ablation resolutions, we reach the same conclusion: map information and deterministic replay contribute complementarily, and no single mechanism can be removed. Because each variant is evaluated against the full \model system, the observed performance drops do not partition the overall gain; instead, multiple components independently account for improvements beyond the VLM-only baseline gap, indicating that each mechanism is load-bearing.

\subsection{Deep Analysis of Map Scale}
\label{sec:mapscale}

\begin{wrapfigure}{r}{0.47\textwidth}
\centering
\vspace{-0.28in}
\includegraphics[width=\linewidth]{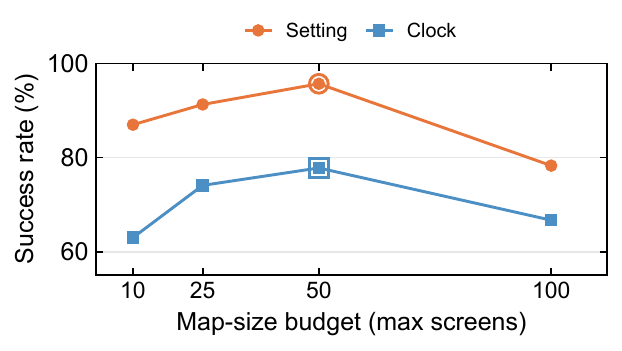}
\vspace{-0.25in}
\caption{Success rate versus the map-size budget (max screens) on two apps (Qwen3.7-Plus).}
\vspace{-0.3in}
\label{fig:mapscale}
\end{wrapfigure}

The ablation above removes components; another complementary question is \emph{how much to build}: what budget should the offline crawler be given, and is a larger map always better? We vary the screen budget of map construction on two representative apps --- Setting (native Android UI) and Clock (moderate complexity) --- and report success rates in Figure~\ref{fig:mapscale}.

Two patterns stand out. \emph{First}, both apps exhibit an inverted-U: success rises as the budget grows from 10 to 50 screens, then drops at 100. Setting climbs from 87.0\% to 95.7\% and falls to 78.3\%; Clock climbs from 63.0\% to 77.8\% and falls to 66.7\%. \emph{Second}, the two ends of the curve fail for different reasons. A small map under-covers the app: more tasks find no matching command and fall back to the VLM, forfeiting the compiled layer. An oversized map inflates the command catalog --- from roughly 130 tokens at 10 screens to 3.5k at 100 --- so Phase~I must choose among a far larger, noisier candidate set, and mis-routing becomes more likely: the compiled layer then actively hurts. The default budget of 50 lands on the peak, which is also why the sweet spot is app-dependent rather than universal. Compression is thus not merely a cost measure but a prerequisite for routing accuracy.

\subsection{Error Analysis}
\label{sec:errors}

\begin{wrapfigure}{r}{0.45\textwidth}
\centering
\vspace{-0.15in}
\includegraphics[width=\linewidth]{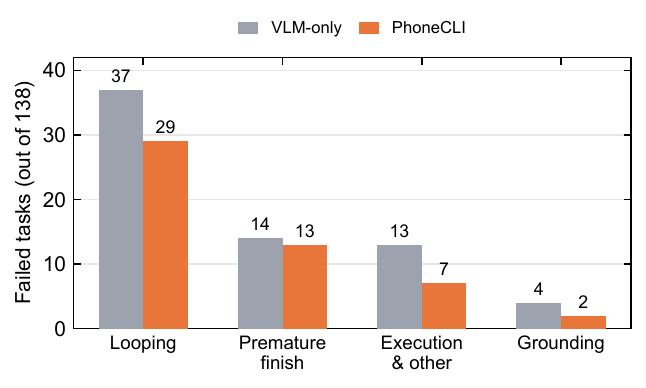}
\vspace{-0.25in}
\caption{Failure-mode counts of the two variants (with Qwen3.7-Plus, on AndroidLab).}
\vspace{-0.1in}
\label{fig:errors}
\end{wrapfigure}

To see what the compiled layer actually fixes, we categorize failed trajectories of the two variants (VLM-only and \model) into four failure modes, cross-checked by deterministic rules where possible (repeated action sequences, aborted trajectories). Figure~\ref{fig:errors} reports the per-mode totals.

The net drop of 17 failures (68 $\rightarrow$ 51) splits into 27 saved and 10 lost. The losses are mostly routing cost --- two tasks mis-judged as command-only and four that landed on mismatched screens or failed the form interaction afterwards --- plus four containing no macro at all, which are single-run VLM variance. The saved 27 span all four modes --- looping 16, premature finish 1, execution 6, grounding 4 --- which is the key finding: the compiled layer repairs no single failure mode, it \emph{prevents the search for the entry point}. Once a command lands the VLM on the right screen, whole classes of failure lose their occasion. Grounding is the cleanest case: all four VLM-only grounding failures are saved, and the two that remain under \model are single-run variance rather than mis-routing, since replay neither taps the wrong element nor is misled by semantics. Execution and other failures lose six, driven by fewer abnormal aborts --- deterministic replay does not crash the way step-by-step interaction does --- and fewer pattern-less exhaustions, as macros save rounds. 

\subsection{Case Study}
\label{sec:case}

\begin{figure}[h]
\centering
\includegraphics[width=0.9\columnwidth]{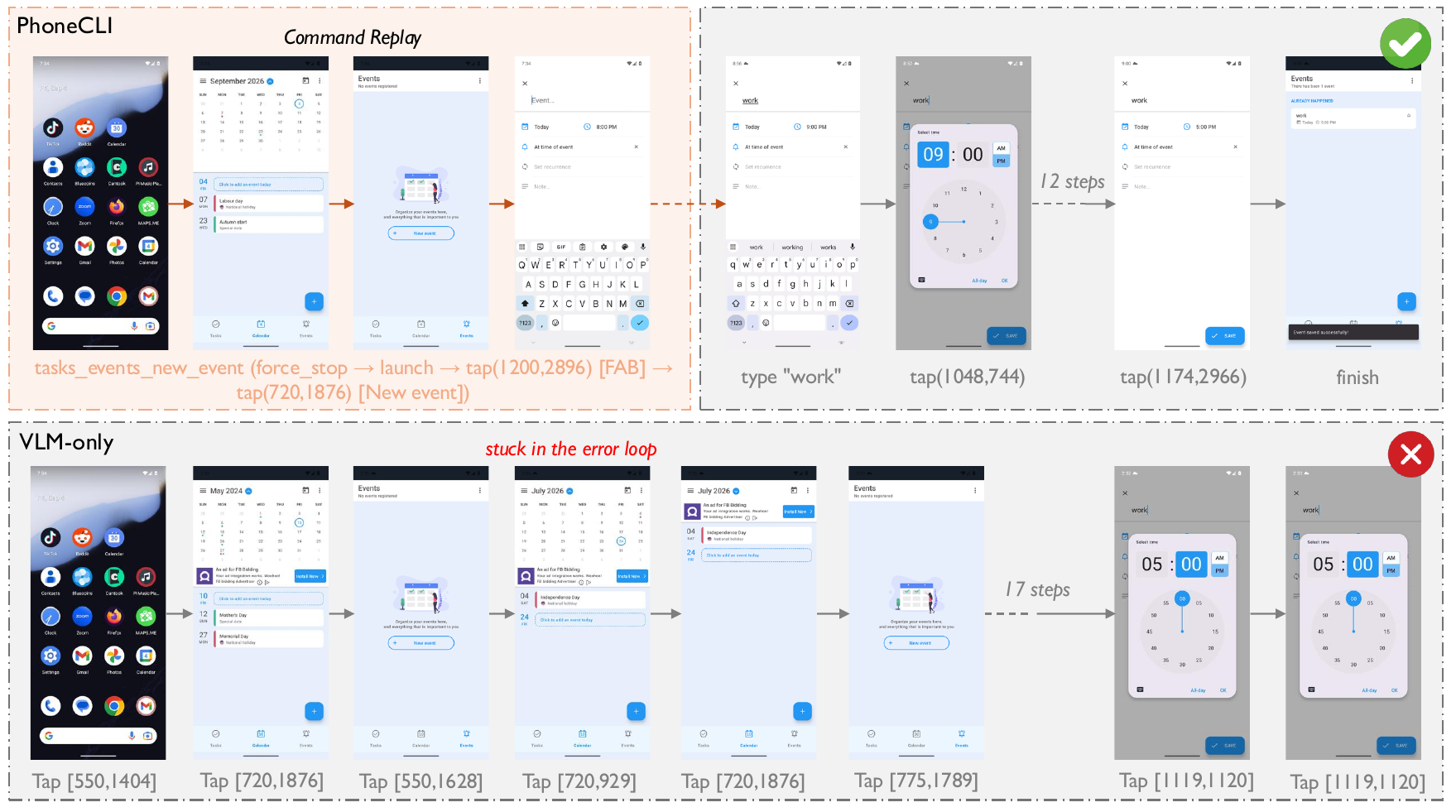}
\caption{Case study on Calendar (``add a 5PM event titled work'').}
\label{fig:case}
\end{figure}

Figure~\ref{fig:case} shows the same task --- add a 5PM event titled ``work'' in Calendar --- under the two conditions. \textbf{\model} (top): routing selects the \texttt{tasks\_events\_new\_event} command, and its replay --- force-stop, launch, tap the floating action button, tap ``new event'' --- lands directly on the event form; the VLM then fills the title (type ``work''), selects the time (\texttt{tap(1048,744)}), and saves (\texttt{tap(1174,2966)}), finishing the task in 17 rounds. \textbf{VLM-only} (bottom): at the same navigation stage that the command replay completes in one shot, the agent gets stuck in an error loop early on, and fails after 25 rounds. 
The comparison isolates the exact failure: task \emph{understanding} was not the bottleneck --- both agents know a 5PM ``work'' event must be created; \emph{locating the entry point} was. Three recorded taps replace lots of exploratory ones, and the compilation pays off.
\section{Related Work}
\label{app:relate}

\subsection{Mobile GUI Agents and Programmatic Execution}

GUI agents for mobile devices largely adopt the perception-action paradigm: each step observes the screen and emits an indexed or coordinate-level action, with the underlying competence acquired through training. One line of work trains vision-language backbones for GUI \emph{grounding} --- perceiving screens and localizing elements \cite{hong2024cogagent, cheng2024seeclick}; another specializes models for \emph{task completion} through supervised fine-tuning or reinforcement learning, including AutoGLM~\cite{liu2024autoglm}, MobileRL~\cite{xu2026mobilerl}, MobileUse~\cite{li2025mobileuse}, UI-Genie-Agent~\cite{xiao2026ui}, UI-Tars~\cite{qin2025ui}, and V-Droid~\cite{dai2025advancing}.  Surveys consolidate this paradigm~\cite{zhang2024large, liu2025llm}. Because capability lives in the weights, a new app demands new data or re-training --- a bottleneck studied directly in recent work on RL generalization~\cite{gu2026generalization} and unseen-app adaptation~\cite{guo2026coadapt}. Orthogonal to this, a second line augments the GUI loop with programmatic execution: agents that write and execute code~\cite{song2026coact}, API-based computer use over instrumented applications~\cite{yan2025mcpworld}, and mobile harnesses that mix GUI with device-side commands~\cite{li2026phoneharness}. All of these assume an execution surface already exists --- an OS that runs code, apps that expose APIs, or system-level CLI access. \model takes the opposite stance: it \emph{synthesizes} that surface by compiling an app's GUI navigation into callable commands, needing no API, no instrumentation, and no training, while keeping the perception-action loop only as a fallback.

\subsection{App Exploration and Structure Mining}

The idea of systematically traversing an app to learn its structure goes back to Android GUI testing crawlers~\cite{su2017guided, gu2019practical}, which optimize coverage and crash discovery rather than agent performance. Recent work reorients exploration toward agents, and diverges by what the explored structure becomes. In the first line, exploration yields \emph{knowledge to read}: AppAgent~\cite{zhang2025appagent} writes natural-language operation documents; AutoDroid~\cite{wen2024autodroid} turns UI transitions into executable scripts; GUI-explorer~\cite{xie2025gui} mines function-aware trajectories; GUI-Xplore~\cite{sun2025gui} packages exploration videos into a generalization dataset; and RAG-GUI~\cite{xu2025retrieval} retrieves web tutorials at inference time. In every case the model consumes the knowledge, yet still performs each action itself. In the second line, exploration yields \emph{graph guidance}: UI-KOBE~\cite{chai2026ui} builds an app knowledge graph offline and lets a lightweight agent navigate by matching its current screen to graph nodes and choosing among their transitions --- the agent stays in the loop, steered but not replaced. In the third line, exploration yields \emph{replay programs}: PreAct~\cite{li2026preact} compiles a successful agent run into a state-machine program online, so that repeating the same task replays instead of re-reasoning. 

\model differs from all three lines on the same axes: it compiles offline, before any task arrives; at the granularity of one navigation command per screen rather than whole task runs; and executes deterministically rather than reading or following. PreAct compiles what an agent has done; \model compiles what the app is. Taken together, the familiar elements \model reuses --- systematic traversal, exploration for knowledge, programmatic execution --- are united by a single decision none of its predecessors makes: compiling the explored structure offline into deterministic commands. 
\section{Conclusion}
\label{sec:conclusion}
\vspace{-0.05in}

In this work, we propose \model to address the lack of mobile app programmatic interfaces. We compile the static, repetitive portion of navigation offline into deterministic, zero-VLM commands and replay the matched command online in sub-second time; open-ended interaction and any compilation failures fall back to the embedded VLM interpreter. On AndroidLab, \model improves task success rate while reducing step and token costs, yielding a stronger agent at lower cost.

\clearpage

\bibliographystyle{unsrtnat}
\bibliography{refs}

\clearpage

\makeatletter
\def\PY@reset{\let\PY@it=\relax \let\PY@bf=\relax%
    \let\PY@ul=\relax \let\PY@tc=\relax%
    \let\PY@bc=\relax \let\PY@ff=\relax}
\def\PY@tok#1{\csname PY@tok@#1\endcsname}
\def\PY@toks#1+{\ifx\relax#1\empty\else%
    \PY@tok{#1}\expandafter\PY@toks\fi}
\def\PY@do#1{\PY@bc{\PY@tc{\PY@ul{%
    \PY@it{\PY@bf{\PY@ff{#1}}}}}}}
\def\PY#1#2{\PY@reset\PY@toks#1+\relax+\PY@do{#2}}

\@namedef{PY@tok@w}{\def\PY@tc##1{\textcolor[rgb]{0.73,0.73,0.73}{##1}}}
\@namedef{PY@tok@c}{\let\PY@it=\textit\def\PY@tc##1{\textcolor[rgb]{0.24,0.48,0.48}{##1}}}
\@namedef{PY@tok@cp}{\def\PY@tc##1{\textcolor[rgb]{0.61,0.40,0.00}{##1}}}
\@namedef{PY@tok@k}{\let\PY@bf=\textbf\def\PY@tc##1{\textcolor[rgb]{0.00,0.50,0.00}{##1}}}
\@namedef{PY@tok@kp}{\def\PY@tc##1{\textcolor[rgb]{0.00,0.50,0.00}{##1}}}
\@namedef{PY@tok@kt}{\def\PY@tc##1{\textcolor[rgb]{0.69,0.00,0.25}{##1}}}
\@namedef{PY@tok@o}{\def\PY@tc##1{\textcolor[rgb]{0.40,0.40,0.40}{##1}}}
\@namedef{PY@tok@ow}{\let\PY@bf=\textbf\def\PY@tc##1{\textcolor[rgb]{0.67,0.13,1.00}{##1}}}
\@namedef{PY@tok@nb}{\def\PY@tc##1{\textcolor[rgb]{0.00,0.50,0.00}{##1}}}
\@namedef{PY@tok@nf}{\def\PY@tc##1{\textcolor[rgb]{0.00,0.00,1.00}{##1}}}
\@namedef{PY@tok@nc}{\let\PY@bf=\textbf\def\PY@tc##1{\textcolor[rgb]{0.00,0.00,1.00}{##1}}}
\@namedef{PY@tok@nn}{\let\PY@bf=\textbf\def\PY@tc##1{\textcolor[rgb]{0.00,0.00,1.00}{##1}}}
\@namedef{PY@tok@ne}{\let\PY@bf=\textbf\def\PY@tc##1{\textcolor[rgb]{0.80,0.25,0.22}{##1}}}
\@namedef{PY@tok@nv}{\def\PY@tc##1{\textcolor[rgb]{0.10,0.09,0.49}{##1}}}
\@namedef{PY@tok@no}{\def\PY@tc##1{\textcolor[rgb]{0.53,0.00,0.00}{##1}}}
\@namedef{PY@tok@nl}{\def\PY@tc##1{\textcolor[rgb]{0.46,0.46,0.00}{##1}}}
\@namedef{PY@tok@ni}{\let\PY@bf=\textbf\def\PY@tc##1{\textcolor[rgb]{0.44,0.44,0.44}{##1}}}
\@namedef{PY@tok@na}{\def\PY@tc##1{\textcolor[rgb]{0.41,0.47,0.13}{##1}}}
\@namedef{PY@tok@nt}{\let\PY@bf=\textbf\def\PY@tc##1{\textcolor[rgb]{0.00,0.50,0.00}{##1}}}
\@namedef{PY@tok@nd}{\def\PY@tc##1{\textcolor[rgb]{0.67,0.13,1.00}{##1}}}
\@namedef{PY@tok@s}{\def\PY@tc##1{\textcolor[rgb]{0.73,0.13,0.13}{##1}}}
\@namedef{PY@tok@sd}{\let\PY@it=\textit\def\PY@tc##1{\textcolor[rgb]{0.73,0.13,0.13}{##1}}}
\@namedef{PY@tok@si}{\let\PY@bf=\textbf\def\PY@tc##1{\textcolor[rgb]{0.64,0.35,0.47}{##1}}}
\@namedef{PY@tok@se}{\let\PY@bf=\textbf\def\PY@tc##1{\textcolor[rgb]{0.67,0.36,0.12}{##1}}}
\@namedef{PY@tok@sr}{\def\PY@tc##1{\textcolor[rgb]{0.64,0.35,0.47}{##1}}}
\@namedef{PY@tok@ss}{\def\PY@tc##1{\textcolor[rgb]{0.10,0.09,0.49}{##1}}}
\@namedef{PY@tok@sx}{\def\PY@tc##1{\textcolor[rgb]{0.00,0.50,0.00}{##1}}}
\@namedef{PY@tok@m}{\def\PY@tc##1{\textcolor[rgb]{0.40,0.40,0.40}{##1}}}
\@namedef{PY@tok@gh}{\let\PY@bf=\textbf\def\PY@tc##1{\textcolor[rgb]{0.00,0.00,0.50}{##1}}}
\@namedef{PY@tok@gu}{\let\PY@bf=\textbf\def\PY@tc##1{\textcolor[rgb]{0.50,0.00,0.50}{##1}}}
\@namedef{PY@tok@gd}{\def\PY@tc##1{\textcolor[rgb]{0.63,0.00,0.00}{##1}}}
\@namedef{PY@tok@gi}{\def\PY@tc##1{\textcolor[rgb]{0.00,0.52,0.00}{##1}}}
\@namedef{PY@tok@gr}{\def\PY@tc##1{\textcolor[rgb]{0.89,0.00,0.00}{##1}}}
\@namedef{PY@tok@ge}{\let\PY@it=\textit}
\@namedef{PY@tok@gs}{\let\PY@bf=\textbf}
\@namedef{PY@tok@gp}{\let\PY@bf=\textbf\def\PY@tc##1{\textcolor[rgb]{0.00,0.00,0.50}{##1}}}
\@namedef{PY@tok@go}{\def\PY@tc##1{\textcolor[rgb]{0.44,0.44,0.44}{##1}}}
\@namedef{PY@tok@gt}{\def\PY@tc##1{\textcolor[rgb]{0.00,0.27,0.87}{##1}}}
\@namedef{PY@tok@err}{\def\PY@bc##1{{\setlength{\fboxsep}{\string -\fboxrule}\fcolorbox[rgb]{1.00,0.00,0.00}{1,1,1}{\strut ##1}}}}
\@namedef{PY@tok@kc}{\let\PY@bf=\textbf\def\PY@tc##1{\textcolor[rgb]{0.00,0.50,0.00}{##1}}}
\@namedef{PY@tok@kd}{\let\PY@bf=\textbf\def\PY@tc##1{\textcolor[rgb]{0.00,0.50,0.00}{##1}}}
\@namedef{PY@tok@kn}{\let\PY@bf=\textbf\def\PY@tc##1{\textcolor[rgb]{0.00,0.50,0.00}{##1}}}
\@namedef{PY@tok@kr}{\let\PY@bf=\textbf\def\PY@tc##1{\textcolor[rgb]{0.00,0.50,0.00}{##1}}}
\@namedef{PY@tok@bp}{\def\PY@tc##1{\textcolor[rgb]{0.00,0.50,0.00}{##1}}}
\@namedef{PY@tok@fm}{\def\PY@tc##1{\textcolor[rgb]{0.00,0.00,1.00}{##1}}}
\@namedef{PY@tok@vc}{\def\PY@tc##1{\textcolor[rgb]{0.10,0.09,0.49}{##1}}}
\@namedef{PY@tok@vg}{\def\PY@tc##1{\textcolor[rgb]{0.10,0.09,0.49}{##1}}}
\@namedef{PY@tok@vi}{\def\PY@tc##1{\textcolor[rgb]{0.10,0.09,0.49}{##1}}}
\@namedef{PY@tok@vm}{\def\PY@tc##1{\textcolor[rgb]{0.10,0.09,0.49}{##1}}}
\@namedef{PY@tok@sa}{\def\PY@tc##1{\textcolor[rgb]{0.73,0.13,0.13}{##1}}}
\@namedef{PY@tok@sb}{\def\PY@tc##1{\textcolor[rgb]{0.73,0.13,0.13}{##1}}}
\@namedef{PY@tok@sc}{\def\PY@tc##1{\textcolor[rgb]{0.73,0.13,0.13}{##1}}}
\@namedef{PY@tok@dl}{\def\PY@tc##1{\textcolor[rgb]{0.73,0.13,0.13}{##1}}}
\@namedef{PY@tok@s2}{\def\PY@tc##1{\textcolor[rgb]{0.73,0.13,0.13}{##1}}}
\@namedef{PY@tok@sh}{\def\PY@tc##1{\textcolor[rgb]{0.73,0.13,0.13}{##1}}}
\@namedef{PY@tok@s1}{\def\PY@tc##1{\textcolor[rgb]{0.73,0.13,0.13}{##1}}}
\@namedef{PY@tok@mb}{\def\PY@tc##1{\textcolor[rgb]{0.40,0.40,0.40}{##1}}}
\@namedef{PY@tok@mf}{\def\PY@tc##1{\textcolor[rgb]{0.40,0.40,0.40}{##1}}}
\@namedef{PY@tok@mh}{\def\PY@tc##1{\textcolor[rgb]{0.40,0.40,0.40}{##1}}}
\@namedef{PY@tok@mi}{\def\PY@tc##1{\textcolor[rgb]{0.40,0.40,0.40}{##1}}}
\@namedef{PY@tok@il}{\def\PY@tc##1{\textcolor[rgb]{0.40,0.40,0.40}{##1}}}
\@namedef{PY@tok@mo}{\def\PY@tc##1{\textcolor[rgb]{0.40,0.40,0.40}{##1}}}
\@namedef{PY@tok@ch}{\let\PY@it=\textit\def\PY@tc##1{\textcolor[rgb]{0.24,0.48,0.48}{##1}}}
\@namedef{PY@tok@cm}{\let\PY@it=\textit\def\PY@tc##1{\textcolor[rgb]{0.24,0.48,0.48}{##1}}}
\@namedef{PY@tok@cpf}{\let\PY@it=\textit\def\PY@tc##1{\textcolor[rgb]{0.24,0.48,0.48}{##1}}}
\@namedef{PY@tok@c1}{\let\PY@it=\textit\def\PY@tc##1{\textcolor[rgb]{0.24,0.48,0.48}{##1}}}
\@namedef{PY@tok@cs}{\let\PY@it=\textit\def\PY@tc##1{\textcolor[rgb]{0.24,0.48,0.48}{##1}}}

\def\PYZbs{\char`\\}
\def\PYZus{\char`\_}
\def\PYZob{\char`\{}
\def\PYZcb{\char`\}}
\def\PYZca{\char`\^}
\def\PYZam{\char`\&}
\def\PYZlt{\char`\<}
\def\PYZgt{\char`\>}
\def\PYZsh{\char`\#}
\def\PYZpc{\char`\%}
\def\PYZdl{\char`\$}
\def\PYZhy{\char`\-}
\def\PYZsq{\char`\'}
\def\PYZdq{\char`\"}
\def\PYZti{\char`\~}
\def\PYZat{@}
\def\PYZlb{[}
\def\PYZrb{]}
\makeatother


\appendix
\section{Appendix}

\subsection{Experimental Setup}
\label{app:setup}

\paragraph{App maps.}
Our maps are built offline, once per app, and reused across every task and every condition. Starting from the app's home screen, a breadth-first crawler explores up to 50 screens to a depth of 3, scrolling each list up to three pages; the only app-specific input is the package name, so no per-app engineering is involved. Each discovered screen is annotated with the text, normalized center, and semantic aliases of its elements, together with the navigation edges it exposes. Across the nine AndroidLab apps the resulting maps contain 320 screens, 2,822 elements, and 1,089 navigation edges, which compile into 2,017 operations (891 navigation and 1,126 action). Building one map takes on the order of ten minutes on a single emulator and runs entirely offline, consuming no rounds or tokens of any evaluated agent. Crucially, construction observes only the app itself: it never reads task instructions, target states, or evaluation code. All variants share the same maps, and the \model w/o replay variant receives the same map serialized as a navigation catalog, without macro replay.

\paragraph{Evaluation.}
Operation tasks are judged by programmatic XML checkers and query tasks by an LLM judge; all conditions are re-judged under one unified judge model (Qwen3.7-Plus) to keep cross-model comparisons free of judge bias. Of the 138 tasks, 94 are graded by deterministic checkers over the final screen XML and the remaining 44 are query-style tasks graded by an LLM judge. Because different agents terminate in different states, we re-grade the logs of every condition with a single unified judge, so that the reported differences reflect the agents rather than the metric. All runs use an Android~13 emulator (Pixel~7~Pro AVD, $1440\times3120$); the emulator is cold-booted once and a clean snapshot is restored before every task, so tasks never observe each other's state. Each task is attempted once under a cap of 25 rounds. Success rate is computed over all 138 tasks, whereas step and token counts are averaged over each condition's successful tasks and therefore measure the cost of a successful trajectory. The gains of our methods over the strongest baseline in Table~\ref{tab:main} and Table~\ref{tab:sota} are statistically significant ($p<0.05$).

\paragraph{Implementation details.}
Observations follow the set-of-mark convention~\cite{yang2023set}: interactable elements are drawn with numeric tags, and the model acts on those tags. All backbones are used off the shelf without fine-tuning, at temperature 0 and at most 512 new tokens per call; Qwen3.7-Plus is our default, and we additionally report runs with GLM-4.6V and Kimi-K3. Every condition shares the same observation format, prompt template, action space, and backbone, and conditions differ only in what is injected into the model's context, which isolates the effect of compiled commands. Executing a compiled operation issues ADB commands directly and invokes no model call. Token usage is accounted by call labels: \texttt{agent\_vlm} for the interpreter loop of Section~\ref{sec:stage3}, \texttt{macro\_map\_task} for the Phase~I routing call of Section~\ref{sec:stage2}, and \texttt{macro\_verify} for its Phase~II verification; judge calls are excluded.

\subsection{App Map Data Structure}
\label{app:map}

The app map is stored as a YAML file. Figure~\ref{fig:map} shows a
simplified excerpt from the Settings app map: each screen records its
elements with normalized centers and navigation edges, and each screen is
associated with a deterministic replay command that reaches it.

\begin{figure}[!ht]
\centering
\begin{tcolorbox}[enhanced, colback=gray!3, colframe=black!30,
                  arc=2mm, boxrule=0.6pt, width=\textwidth,
                  title={\small App-map excerpt (Settings, simplified)}]
\begin{Verbatim}[commandchars=\\\{\},fontsize=\small,breaklines=true]
\PY{n+nt}{screens}\PY{p}{:}
\PY{p+pIndicator}{\PYZhy{}}\PY{+w}{ }\PY{n+nt}{id}\PY{p}{:}\PY{+w}{ }\PY{l+lScalar+lScalarPlain}{screen\PYZus{}0}\PY{+w}{                          }\PY{c+c1}{\PYZsh{} Settings home screen}
\PY{+w}{  }\PY{n+nt}{description}\PY{p}{:}\PY{+w}{ }\PY{l+s}{\PYZdq{}}\PY{l+s}{Main}\PY{n+nv}{ }\PY{l+s}{Settings}\PY{n+nv}{ }\PY{l+s}{screen}\PY{n+nv}{ }\PY{l+s}{with}\PY{n+nv}{ }\PY{l+s}{a}\PY{n+nv}{ }\PY{l+s}{profile}\PY{n+nv}{ }\PY{l+s}{header,}\PY{n+nv}{ }\PY{l+s}{search}\PY{n+nv}{ }\PY{l+s}{bar,}\PY{n+nv}{ }\PY{l+s}{and}\PY{n+nv}{ }\PY{l+s}{a}\PY{n+nv}{ }\PY{l+s}{comprehensive}\PY{n+nv}{ }\PY{l+s}{list}\PY{n+nv}{ }\PY{l+s}{o}\PY{n+nv}{ }\PY{l+s}{system}\PY{n+nv}{ }\PY{l+s}{configuration}\PY{n+nv}{ }\PY{l+s}{categories.}\PY{l+s}{\PYZdq{}}
\PY{+w}{  }\PY{n+nt}{elements}\PY{p}{:}
\PY{+w}{  }\PY{p+pIndicator}{\PYZhy{}}\PY{+w}{ }\PY{n+nt}{text}\PY{p}{:}\PY{+w}{ }\PY{l+lScalar+lScalarPlain}{Connected}\PY{l+lScalar+lScalarPlain}{ }\PY{l+lScalar+lScalarPlain}{devices}
\PY{+w}{    }\PY{n+nt}{center}\PY{p}{:}\PY{+w}{ }\PY{p+pIndicator}{[}\PY{n+nv}{0.3802}\PY{p+pIndicator}{,}\PY{+w}{ }\PY{n+nv}{0.4575}\PY{p+pIndicator}{]}\PY{+w}{            }\PY{c+c1}{\PYZsh{} normalized (x, y); resolution\PYZhy{}independent}
\PY{+w}{    }\PY{n+nt}{found\PYZus{}at\PYZus{}scroll}\PY{p}{:}\PY{+w}{ }\PY{l+lScalar+lScalarPlain}{0}\PY{+w}{                  }\PY{c+c1}{\PYZsh{} 0 = visible on the first screenful}
\PY{+w}{    }\PY{n+nt}{leads\PYZus{}to}\PY{p}{:}\PY{+w}{ }\PY{l+lScalar+lScalarPlain}{screen\PYZus{}5}\PY{+w}{                  }\PY{c+c1}{\PYZsh{} navigation edge: tapping this reaches screen\PYZus{}5}
\PY{+w}{    }\PY{n+nt}{aliases}\PY{p}{:}\PY{+w}{ }\PY{p+pIndicator}{[}\PY{n+nv}{devices}\PY{p+pIndicator}{,}\PY{+w}{ }\PY{n+nv}{connections}\PY{p+pIndicator}{]}\PY{+w}{     }\PY{c+c1}{\PYZsh{} LLM\PYZhy{}generated aliases}
\PY{+w}{    }\PY{n+nt}{semantic\PYZus{}type}\PY{p}{:}\PY{+w}{ }\PY{l+lScalar+lScalarPlain}{setting}
\PY{p+pIndicator}{\PYZhy{}}\PY{+w}{ }\PY{n+nt}{id}\PY{p}{:}\PY{+w}{ }\PY{l+lScalar+lScalarPlain}{screen\PYZus{}5}\PY{+w}{                          }\PY{c+c1}{\PYZsh{} Connected devices page (one tap from home)}
\PY{+w}{  }\PY{n+nt}{description}\PY{p}{:}\PY{+w}{ }\PY{l+s}{\PYZdq{}}\PY{l+s}{Connected}\PY{n+nv}{ }\PY{l+s}{devices}\PY{n+nv}{ }\PY{l+s}{settings}\PY{n+nv}{ }\PY{l+s}{page}\PY{n+nv}{ }\PY{l+s}{where}\PY{n+nv}{ }\PY{l+s}{users}\PY{n+nv}{ }\PY{l+s}{can}\PY{n+nv}{ }\PY{l+s}{pair}\PY{n+nv}{ }\PY{l+s}{new}\PY{n+nv}{ }\PY{l+s}{devices,}\PY{n+nv}{ }\PY{l+s}{view}\PY{n+nv}{ }\PY{l+s}{saved}\PY{n+nv}{ }\PY{l+s}{devices,}\PY{n+nv}{ }\PY{l+s}{and}\PY{n+nv}{ }\PY{l+s}{manage}\PY{n+nv}{ }\PY{l+s}{connection}\PY{n+nv}{ }\PY{l+s}{preferences.}\PY{l+s}{\PYZdq{}}
\PY{+w}{  }\PY{n+nt}{elements}\PY{p}{:}
\PY{+w}{  }\PY{p+pIndicator}{\PYZhy{}}\PY{+w}{ }\PY{n+nt}{text}\PY{p}{:}\PY{+w}{ }\PY{l+lScalar+lScalarPlain}{Pair}\PY{l+lScalar+lScalarPlain}{ }\PY{l+lScalar+lScalarPlain}{new}\PY{l+lScalar+lScalarPlain}{ }\PY{l+lScalar+lScalarPlain}{device}
\PY{+w}{    }\PY{n+nt}{center}\PY{p}{:}\PY{+w}{ }\PY{p+pIndicator}{[}\PY{n+nv}{0.3441}\PY{p+pIndicator}{,}\PY{+w}{ }\PY{n+nv}{0.2800}\PY{p+pIndicator}{]}
\PY{+w}{    }\PY{n+nt}{found\PYZus{}at\PYZus{}scroll}\PY{p}{:}\PY{+w}{ }\PY{l+lScalar+lScalarPlain}{0}
\PY{+w}{    }\PY{n+nt}{leads\PYZus{}to}\PY{p}{:}\PY{+w}{ }\PY{l+lScalar+lScalarPlain}{screen\PYZus{}25}
\PY{+w}{    }\PY{n+nt}{aliases}\PY{p}{:}\PY{+w}{ }\PY{p+pIndicator}{[}\PY{n+nv}{add}\PY{n+nv}{ }\PY{n+nv}{device}\PY{p+pIndicator}{,}\PY{+w}{ }\PY{n+nv}{connect}\PY{n+nv}{ }\PY{n+nv}{device}\PY{p+pIndicator}{]}
\PY{+w}{    }\PY{n+nt}{semantic\PYZus{}type}\PY{p}{:}\PY{+w}{ }\PY{l+lScalar+lScalarPlain}{button}
\PY{n+nt}{screen\PYZus{}macros}\PY{p}{:}\PY{+w}{                          }\PY{c+c1}{\PYZsh{} compiled command: clean start \PYZhy{}\PYZgt{} screen\PYZus{}25}
\PY{+w}{  }\PY{n+nt}{screen\PYZus{}25}\PY{p}{:}\PY{+w}{                            }\PY{c+c1}{\PYZsh{} Bluetooth settings page}
\PY{+w}{  }\PY{p+pIndicator}{\PYZhy{}}\PY{+w}{ }\PY{n+nt}{action}\PY{p}{:}\PY{+w}{ }\PY{l+lScalar+lScalarPlain}{force\PYZus{}stop}
\PY{+w}{    }\PY{n+nt}{package}\PY{p}{:}\PY{+w}{ }\PY{l+lScalar+lScalarPlain}{com.android.settings}
\PY{+w}{    }\PY{n+nt}{wait}\PY{p}{:}\PY{+w}{ }\PY{l+lScalar+lScalarPlain}{0.5}
\PY{+w}{  }\PY{p+pIndicator}{\PYZhy{}}\PY{+w}{ }\PY{n+nt}{action}\PY{p}{:}\PY{+w}{ }\PY{l+lScalar+lScalarPlain}{launch}
\PY{+w}{    }\PY{n+nt}{package}\PY{p}{:}\PY{+w}{ }\PY{l+lScalar+lScalarPlain}{com.android.settings}
\PY{+w}{    }\PY{n+nt}{wait}\PY{p}{:}\PY{+w}{ }\PY{l+lScalar+lScalarPlain}{3.0}
\PY{+w}{  }\PY{p+pIndicator}{\PYZhy{}}\PY{+w}{ }\PY{n+nt}{action}\PY{p}{:}\PY{+w}{ }\PY{l+lScalar+lScalarPlain}{tap}
\PY{+w}{    }\PY{n+nt}{x}\PY{p}{:}\PY{+w}{ }\PY{l+lScalar+lScalarPlain}{548}\PY{+w}{                              }\PY{c+c1}{\PYZsh{} taps \PYZdq{}Connected devices\PYZdq{} on screen\PYZus{}0}
\PY{+w}{    }\PY{n+nt}{y}\PY{p}{:}\PY{+w}{ }\PY{l+lScalar+lScalarPlain}{1428}
\PY{+w}{    }\PY{n+nt}{wait}\PY{p}{:}\PY{+w}{ }\PY{l+lScalar+lScalarPlain}{1.5}
\PY{+w}{  }\PY{p+pIndicator}{\PYZhy{}}\PY{+w}{ }\PY{n+nt}{action}\PY{p}{:}\PY{+w}{ }\PY{l+lScalar+lScalarPlain}{tap}
\PY{+w}{    }\PY{n+nt}{x}\PY{p}{:}\PY{+w}{ }\PY{l+lScalar+lScalarPlain}{496}\PY{+w}{                              }\PY{c+c1}{\PYZsh{} taps \PYZdq{}Pair new device\PYZdq{} on screen\PYZus{}5}
\PY{+w}{    }\PY{n+nt}{y}\PY{p}{:}\PY{+w}{ }\PY{l+lScalar+lScalarPlain}{874}
\PY{+w}{    }\PY{n+nt}{wait}\PY{p}{:}\PY{+w}{ }\PY{l+lScalar+lScalarPlain}{1.5}
\end{Verbatim}
\end{tcolorbox}
\caption{App-map excerpt (Settings, simplified): elements, the navigation edges
they define, and the deterministic replay command compiled from those edges.
Tap coordinates equal \texttt{center}$\times$screen size, and invoking a
compiled command costs no VLM call.}
\label{fig:map}
\end{figure}

\subsection{From App Maps to Callable Commands}
\label{app:catalog}

Figure~\ref{fig:map} shows what an app map stores; this section describes the second half of Stage~1, which turns those records into the command catalog $\mathcal{C}$ that the Stage~2 router reads, and illustrates the one-line interface the router answers with.

\textbf{Compiling commands.} The compiler walks the screen graph from the home screen and, for every interactive element, emits one \emph{operation}: the element path that leads to it (for example \texttt{Connected devices} $\rightarrow$ \texttt{Pair new device}) paired with the action sequence that produces it, namely the macro of the parent screen, followed by the scroll steps needed to bring the element into view, followed by a tap at its recorded center. Operations whose element navigates to another screen are marked \textsc{nav}; the remainder are terminal actions (\textsc{act}) such as toggles and text fields. Because several element paths may reach the same screen, an operation is keyed by its \emph{path}, not by its destination: the catalog lists one entry per path, and only the path along which a screen was first expanded carries that screen's own children.

\textbf{Rendering the catalog.}
Because the catalog is re-sent on every routing call, its size is a recurring cost, so we compress it aggressively. The router only has to decide \emph{where} to go, so the catalog lists navigation operations only: for Settings this reduces 288 operations to 98 \textsc{nav} entries in 9 groups, and the rendered catalog from roughly 6.2k to 1.9k tokens. Entries sharing a common path prefix are merged into a tree, so a shared navigation prefix is written once and children are indented. Each line carries the element text, its semantic type, one of its aliases, and a stable operation id --- a slug of its element path, truncated at 60 characters --- which is the only thing the router has to output.

\textbf{From catalog to action.}
The router answers in a single line with one of four decisions (\textsc{op}, \textsc{macro\_vlm}, \textsc{need\_vlm}, \textsc{finish}); the selected id resolves to the macro compiled above, which is replayed with fixed coordinates and no model call. For example, for the task \emph{``pair a new Bluetooth device''} the router answers \texttt{MACRO\_VLM:~settings\_connected\_devices\_pair\_new\_device} and the system replays \texttt{force\_stop} $\mid$ \texttt{launch} $\mid$ \texttt{tap(548,1428)} $\mid$ \texttt{tap(496,874)} before handing control to the VLM on the Bluetooth page. The catalog is compiled once per app and cached, so this interface costs one prefix of a single routing call per task, and it is plain text: it can be inspected, edited, or reused across models.

\begin{figure}[h]
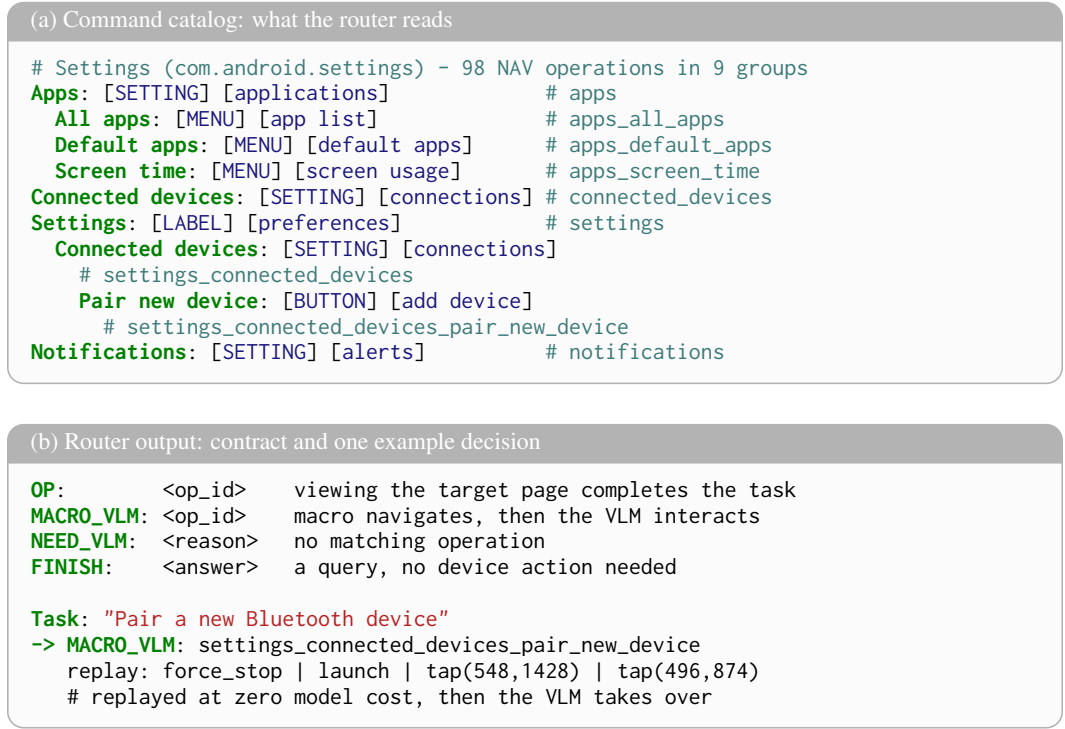

\centering

\begin{tcolorbox}[enhanced, colback=gray!3, colframe=black!30,
                  arc=2mm, boxrule=0.6pt, width=\textwidth,
                  left=2mm, right=2mm, top=1mm, bottom=1mm, boxsep=1mm,
                  title={\small (a) Command catalog: what the router reads}]
\begin{Verbatim}[commandchars=\\\{\},fontsize=\small,breaklines=true]
\PY{c+c1}{\PYZsh{} Settings (com.android.settings) \PYZhy{} 98 NAV operations in 9 groups}
\PY{n+nt}{Apps}\PY{p}{:}\PY{+w}{ }\PY{p+pIndicator}{[}\PY{n+nv}{SETTING}\PY{p+pIndicator}{]}\PY{+w}{ }\PY{p+pIndicator}{[}\PY{n+nv}{applications}\PY{p+pIndicator}{]}\PY{+w}{             }\PY{c+c1}{\PYZsh{} apps}
\PY{+w}{  }\PY{n+nt}{All apps}\PY{p}{:}\PY{+w}{ }\PY{p+pIndicator}{[}\PY{n+nv}{MENU}\PY{p+pIndicator}{]}\PY{+w}{ }\PY{p+pIndicator}{[}\PY{n+nv}{app}\PY{n+nv}{ }\PY{n+nv}{list}\PY{p+pIndicator}{]}\PY{+w}{              }\PY{c+c1}{\PYZsh{} apps\PYZus{}all\PYZus{}apps}
\PY{+w}{  }\PY{n+nt}{Default apps}\PY{p}{:}\PY{+w}{ }\PY{p+pIndicator}{[}\PY{n+nv}{MENU}\PY{p+pIndicator}{]}\PY{+w}{ }\PY{p+pIndicator}{[}\PY{n+nv}{default}\PY{n+nv}{ }\PY{n+nv}{apps}\PY{p+pIndicator}{]}\PY{+w}{      }\PY{c+c1}{\PYZsh{} apps\PYZus{}default\PYZus{}apps}
\PY{+w}{  }\PY{n+nt}{Screen time}\PY{p}{:}\PY{+w}{ }\PY{p+pIndicator}{[}\PY{n+nv}{MENU}\PY{p+pIndicator}{]}\PY{+w}{ }\PY{p+pIndicator}{[}\PY{n+nv}{screen}\PY{n+nv}{ }\PY{n+nv}{usage}\PY{p+pIndicator}{]}\PY{+w}{       }\PY{c+c1}{\PYZsh{} apps\PYZus{}screen\PYZus{}time}
\PY{n+nt}{Connected devices}\PY{p}{:}\PY{+w}{ }\PY{p+pIndicator}{[}\PY{n+nv}{SETTING}\PY{p+pIndicator}{]}\PY{+w}{ }\PY{p+pIndicator}{[}\PY{n+nv}{connections}\PY{p+pIndicator}{]}\PY{+w}{ }\PY{c+c1}{\PYZsh{} connected\PYZus{}devices}
\PY{n+nt}{Settings}\PY{p}{:}\PY{+w}{ }\PY{p+pIndicator}{[}\PY{n+nv}{LABEL}\PY{p+pIndicator}{]}\PY{+w}{ }\PY{p+pIndicator}{[}\PY{n+nv}{preferences}\PY{p+pIndicator}{]}\PY{+w}{            }\PY{c+c1}{\PYZsh{} settings}
\PY{+w}{  }\PY{n+nt}{Connected devices}\PY{p}{:}\PY{+w}{ }\PY{p+pIndicator}{[}\PY{n+nv}{SETTING}\PY{p+pIndicator}{]}\PY{+w}{ }\PY{p+pIndicator}{[}\PY{n+nv}{connections}\PY{p+pIndicator}{]}
\PY{+w}{    }\PY{c+c1}{\PYZsh{} settings\PYZus{}connected\PYZus{}devices}
\PY{+w}{    }\PY{n+nt}{Pair new device}\PY{p}{:}\PY{+w}{ }\PY{p+pIndicator}{[}\PY{n+nv}{BUTTON}\PY{p+pIndicator}{]}\PY{+w}{ }\PY{p+pIndicator}{[}\PY{n+nv}{add}\PY{n+nv}{ }\PY{n+nv}{device}\PY{p+pIndicator}{]}
\PY{+w}{      }\PY{c+c1}{\PYZsh{} settings\PYZus{}connected\PYZus{}devices\PYZus{}pair\PYZus{}new\PYZus{}device}
\PY{n+nt}{Notifications}\PY{p}{:}\PY{+w}{ }\PY{p+pIndicator}{[}\PY{n+nv}{SETTING}\PY{p+pIndicator}{]}\PY{+w}{ }\PY{p+pIndicator}{[}\PY{n+nv}{alerts}\PY{p+pIndicator}{]}\PY{+w}{          }\PY{c+c1}{\PYZsh{} notifications}
\end{Verbatim}
\end{tcolorbox}

\vspace{4pt}

\begin{tcolorbox}[enhanced, colback=gray!3, colframe=black!30,
                  arc=2mm, boxrule=0.6pt, width=\textwidth,
                  left=2mm, right=2mm, top=1mm, bottom=1mm, boxsep=1mm,
                  title={\small (b) Router output: contract and one example decision}]
\begin{Verbatim}[commandchars=\\\{\},fontsize=\small,breaklines=true]
\PY{n+nt}{OP}\PY{p}{:}\PY{+w}{        }\PY{l+lScalar+lScalarPlain}{\PYZlt{}op\PYZus{}id\PYZgt{}}\PY{l+lScalar+lScalarPlain}{    }\PY{l+lScalar+lScalarPlain}{viewing}\PY{l+lScalar+lScalarPlain}{ }\PY{l+lScalar+lScalarPlain}{the}\PY{l+lScalar+lScalarPlain}{ }\PY{l+lScalar+lScalarPlain}{target}\PY{l+lScalar+lScalarPlain}{ }\PY{l+lScalar+lScalarPlain}{page}\PY{l+lScalar+lScalarPlain}{ }\PY{l+lScalar+lScalarPlain}{completes}\PY{l+lScalar+lScalarPlain}{ }\PY{l+lScalar+lScalarPlain}{the}\PY{l+lScalar+lScalarPlain}{ }\PY{l+lScalar+lScalarPlain}{task}
\PY{n+nt}{MACRO\PYZus{}VLM}\PY{p}{:}\PY{+w}{ }\PY{l+lScalar+lScalarPlain}{\PYZlt{}op\PYZus{}id\PYZgt{}}\PY{l+lScalar+lScalarPlain}{    }\PY{l+lScalar+lScalarPlain}{macro}\PY{l+lScalar+lScalarPlain}{ }\PY{l+lScalar+lScalarPlain}{navigates,}\PY{l+lScalar+lScalarPlain}{ }\PY{l+lScalar+lScalarPlain}{then}\PY{l+lScalar+lScalarPlain}{ }\PY{l+lScalar+lScalarPlain}{the}\PY{l+lScalar+lScalarPlain}{ }\PY{l+lScalar+lScalarPlain}{VLM}\PY{l+lScalar+lScalarPlain}{ }\PY{l+lScalar+lScalarPlain}{interacts}
\PY{n+nt}{NEED\PYZus{}VLM}\PY{p}{:}\PY{+w}{  }\PY{l+lScalar+lScalarPlain}{\PYZlt{}reason\PYZgt{}}\PY{l+lScalar+lScalarPlain}{   }\PY{l+lScalar+lScalarPlain}{no}\PY{l+lScalar+lScalarPlain}{ }\PY{l+lScalar+lScalarPlain}{matching}\PY{l+lScalar+lScalarPlain}{ }\PY{l+lScalar+lScalarPlain}{operation}
\PY{n+nt}{FINISH}\PY{p}{:}\PY{+w}{    }\PY{l+lScalar+lScalarPlain}{\PYZlt{}answer\PYZgt{}}\PY{l+lScalar+lScalarPlain}{   }\PY{l+lScalar+lScalarPlain}{a}\PY{l+lScalar+lScalarPlain}{ }\PY{l+lScalar+lScalarPlain}{query,}\PY{l+lScalar+lScalarPlain}{ }\PY{l+lScalar+lScalarPlain}{no}\PY{l+lScalar+lScalarPlain}{ }\PY{l+lScalar+lScalarPlain}{device}\PY{l+lScalar+lScalarPlain}{ }\PY{l+lScalar+lScalarPlain}{action}\PY{l+lScalar+lScalarPlain}{ }\PY{l+lScalar+lScalarPlain}{needed}

\PY{n+nt}{Task}\PY{p}{:}\PY{+w}{ }\PY{l+s}{\PYZdq{}}\PY{l+s}{Pair}\PY{n+nv}{ }\PY{l+s}{a}\PY{n+nv}{ }\PY{l+s}{new}\PY{n+nv}{ }\PY{l+s}{Bluetooth}\PY{n+nv}{ }\PY{l+s}{device}\PY{l+s}{\PYZdq{}}
\PY{n+nt}{\PYZhy{}\PYZgt{} MACRO\PYZus{}VLM}\PY{p}{:}\PY{+w}{ }\PY{l+lScalar+lScalarPlain}{settings\PYZus{}connected\PYZus{}devices\PYZus{}pair\PYZus{}new\PYZus{}device}
\PY{+w}{   }\PY{l+lScalar+lScalarPlain}{replay}\PY{p+pIndicator}{:}\PY{+w}{ }\PY{l+lScalar+lScalarPlain}{force\PYZus{}stop}\PY{l+lScalar+lScalarPlain}{ }\PY{l+lScalar+lScalarPlain}{|}\PY{l+lScalar+lScalarPlain}{ }\PY{l+lScalar+lScalarPlain}{launch}\PY{l+lScalar+lScalarPlain}{ }\PY{l+lScalar+lScalarPlain}{|}\PY{l+lScalar+lScalarPlain}{ }\PY{l+lScalar+lScalarPlain}{tap(548,1428)}\PY{l+lScalar+lScalarPlain}{ }\PY{l+lScalar+lScalarPlain}{|}\PY{l+lScalar+lScalarPlain}{ }\PY{l+lScalar+lScalarPlain}{tap(496,874)}
\PY{+w}{   }\PY{l+lScalar+lScalarPlain}{\PYZsh{}}\PY{l+lScalar+lScalarPlain}{ }\PY{l+lScalar+lScalarPlain}{replayed}\PY{l+lScalar+lScalarPlain}{ }\PY{l+lScalar+lScalarPlain}{at}\PY{l+lScalar+lScalarPlain}{ }\PY{l+lScalar+lScalarPlain}{zero}\PY{l+lScalar+lScalarPlain}{ }\PY{l+lScalar+lScalarPlain}{model}\PY{l+lScalar+lScalarPlain}{ }\PY{l+lScalar+lScalarPlain}{cost,}\PY{l+lScalar+lScalarPlain}{ }\PY{l+lScalar+lScalarPlain}{then}\PY{l+lScalar+lScalarPlain}{ }\PY{l+lScalar+lScalarPlain}{the}\PY{l+lScalar+lScalarPlain}{ }\PY{l+lScalar+lScalarPlain}{VLM}\PY{l+lScalar+lScalarPlain}{ }\PY{l+lScalar+lScalarPlain}{takes}\PY{l+lScalar+lScalarPlain}{ }\PY{l+lScalar+lScalarPlain}{over}
\end{Verbatim}
\end{tcolorbox}

\caption{Command catalog compiled from the app map of Figure~\ref{fig:map}.}
\label{fig:catalog}
\end{figure}

\subsection{Generalization to AndroidWorld}
\label{sec:androidworld}

\model is an enhancement layer, and the strongest test of that claim is whether it helps an agent it was not designed for. We therefore attach the compiled layer to M3A --- the Multimodal Autonomous Agent and official baseline of AndroidWorld~\cite{rawles2025androidworld}, which acts step by step on annotated screenshots --- by subclassing it and adding only the map-based routing of Section~\ref{sec:stage2}. M3A's prompts, action space, and perception remain untouched, and when routing declines, execution proceeds exactly as the original M3A. Both agents run on Qwen3.7-Plus.

\begin{wraptable}{r}{0.46\textwidth}
\vspace{-1.2em}
\centering
\caption{AndroidWorld results.}
\footnotesize
\begin{tabular}{@{}l c c c@{}}
\toprule
Agent & Success & Tokens & Steps \\
\midrule
M3A (official) & 61.2\% & 27.58M & 8.4 \\
\textbf{M3A w \model} & \textbf{65.5\%} & \textbf{25.04M} & \textbf{7.7} \\
\bottomrule
\end{tabular}
\label{tab:aworld}
\end{wraptable}

Table~\ref{tab:aworld} reports the results. The map-augmented agent scores higher than the original M3A agent. And the efficiency gains are consistent and substantial: 9\% fewer tokens overall, and on the 70 tasks both agents solve, 7.7 steps versus 8.4. That these savings transfer to a different benchmark, a different agent framework, and an untouched M3A shows the compiled layer's efficiency does not depend on any AndroidLab-specific design. M3A is already a mature agent, and on this backbone the enhancement approaches the ceiling of what Qwen3.7-Plus can achieve --- what remains to gain is efficiency rather than capability. Across environments, the enhancement-layer claim of our method therefore holds.

\end{document}